\documentclass[11pt]{article}

\usepackage[final]{acl}

\usepackage{times}
\usepackage{latexsym}
\usepackage{amsmath}
\usepackage{enumitem}
\usepackage{booktabs} 
\usepackage{algorithm}
\usepackage{algpseudocode}
\usepackage{listings}
\usepackage[T1]{fontenc}

\usepackage[utf8]{inputenc}

\usepackage{microtype}

\usepackage{inconsolata}

\usepackage{graphicx}

\title{SKILL.state: Scalable Long-Horizon Agent Skills}

\author{
 \textbf{Sanket Badhe\textsuperscript{1}},
 \textbf{Priyanka Tiwari\textsuperscript{2},
  \textbf{Jonghyun Chung\textsuperscript{1}},}
\\
\\
 \textsuperscript{1}Google LLC,
 \textsuperscript{2}Purdue University
    \textbf{Correspondence:} \href{mailto:sanketbadhe@google.com}{sanketbadhe@google.com}
}

\begin{document}
\maketitle
\begin{abstract}
Large Language Models (LLMs) increasingly act as autonomous agents executing complex, long-running procedural skills. Existing agent runtimes maintain execution by continually appending observations, actions, and intermediate reasoning traces to an ever-growing conversation history, causing latency degradation and context-poisoning failures over long horizons. We present SKILL.state, a runtime architecture that replaces append-only conversational history with an explicit, mutable execution state. At each execution step, the model receives only the immutable skill specification, the current structured execution state, and the latest observation. Intermediate reasoning is discarded immediately after producing a validated state update, preventing prompt growth with execution history. Across diverse datasets, models, and execution environments, SKILL.state improves task accuracy while substantially reducing cumulative token consumption. Our results demonstrate that explicit execution state is an effective and architecture-agnostic abstraction for scalable long-horizon agent skills.
\end{abstract}

\section{Introduction}
\label{sec:intro}

Large Language Models (LLMs) have rapidly evolved from passive language interfaces into autonomous systems capable of iterative reasoning, tool use, and interaction with external environments \citep{yao2022react, schick2023toolformer, qin2023toolllm, wu2023autogen}. Recent work further demonstrates that these capabilities can be encapsulated as reusable procedural skills, enabling agents to perform software engineering, workflow automation, web interaction, and scientific discovery through modular compositions of specialized behaviors \citep{badhe2026systematic}. As agents increasingly execute long-running procedures, execution itself becomes a systems problem rather than purely a reasoning problem.

Modern agent runtimes almost universally adopt a conversational execution model. At every execution step, the language model receives the original skill specification together with an ever-growing transcript of previous reasoning, actions, observations, and tool outputs \citep{yao2022react, mialon2023augmented}. Although memory systems alleviate context growth through summarization or retrieval \citep{packer2023memgpt, xu2023longllmem, zhong2023memorybank}, they preserve the same execution semantics: future decisions are conditioned on textual reconstructions of past execution rather than an explicit representation of the current execution state.

This design introduces fundamental limitations for long-horizon procedural skills. Prompt size grows with execution length, increasing token consumption and inference cost \citep{liu2024lost, xiao2023streamingllm}. Historical observations and obsolete reasoning remain embedded in the context long after they cease to be relevant, requiring the model to continually distinguish current facts from historical artifacts. Consequently, execution correctness increasingly depends on reconstructing state from accumulated textual history.

In this paper, we introduce SKILL.state, a runtime architecture that reformulates procedural skill execution as explicit state transitions rather than conversational history accumulation. Figure~\ref{fig:overview} provides an overview of the proposed runtime. At execution step $t$, the language model receives only three inputs:
\begin{equation}
A_t = (P,\Sigma_t,O_t),
\end{equation}
where $P$ denotes the immutable procedural specification, $\Sigma_t$ is the structured execution state, and $O_t$ is the latest environment observation. After producing a validated state update, the intermediate reasoning trace is discarded while only the updated execution state is retained. Consequently, execution depends strictly on the current world state instead of replaying historical trajectories.

To evaluate this hypothesis, we evaluate SKILL.state across both synthetic and real-world benchmarks: SkillExecBench, a controlled benchmark designed for long-horizon procedural skill execution under scaling, noise, and state recovery; InterCode CTF \citep{yang2023intercode}, featuring interactive Linux terminal exploitation; and Sierra $\tau$-Bench \citep{yao2024tau}, evaluating multi-turn customer-service workflows over complex database APIs.

Experimental results demonstrate that explicit execution state substantially improves the scalability of long-horizon procedural skills by maintaining bounded prompt sizes while cutting token consumption and outperforming history-based and compression-based baselines across multiple model families.

Our contributions are summarized as follows:
\begin{itemize}
    \item We propose SKILL.state, a runtime architecture that executes procedural skills through explicit structured execution state where intermediate reasoning is discarded after each step, proving a strictly bounded $\mathcal{O}(1)$ prompt footprint and $\mathcal{O}(T)$ cumulative token complexity.
    \item We present SkillExecBench, alongside evaluations on public benchmarks (InterCode CTF and Sierra $\tau$-Bench), for evaluating long-horizon procedural skill execution in sequential, stateful environments.
    \item Across multiple execution horizons and runtime baselines, we demonstrate that state-centric execution maintains competitive task performance while substantially reducing prompt growth and cumulative token consumption across both proprietary and open-weight models.
\end{itemize}

\section{Related Work}

\subsection{Procedural Skills for LLM Agents}
Existing research on reusable procedural skills primarily addresses skill discovery, representation, composition and and security threat modeling \citep{badhe2026systematic, badhe2026agentskillsecuritythreat}. Our work instead focuses on the largely unexplored mechanics of skill execution once a skill has been selected.

\subsection{Memory Architectures for Long-Horizon Agents}
Long-horizon agent architectures typically preserve conversational semantics through episodic retrieval \citep{park2023generative} or persistent storage \citep{chhikara2025mem0, zhong2024memorybank}. These methods leave execution state implicitly distributed across accumulated logs. SKILL.state instead isolates execution into an explicit, mutable runtime state, eliminating the need to repeatedly reconstruct world models from textual history. Frameworks like LangGraph use auxiliary structured state to orchestrate workflows across agent nodes. However, these systems still rely on conversational transcripts as the primary reasoning substrate. SKILL.state replaces this substrate by discarding intermediate reasoning traces immediately after producing validated state transitions.

\subsection{Dialogue State Tracking}
Dialogue State Tracking (DST) maintains user slot values across conversational turns in task-oriented dialogue \citep{williams2013dialog, henderson2014second, rastogi2020towards, wu2019transferable, heck2020trippy, hosseini2020simple}. While both DST and SKILL.state maintain structured representations, they differ fundamentally in execution mechanics: DST tracks auxiliary state alongside full conversational transcripts in quasi-static dialogues, whereas SKILL.state treats the structured state as a sufficient statistic, discarding conversational history to execute autonomous skills in dynamic environments with bounded prompt footprints.

\subsection{Context Management and Long-Context Reasoning}
Language models exhibit degraded retrieval over long contexts \citep{liu2024lost, zhang2024bench}, motivating streaming attention \citep{xiao2024streamingllm} and prompt compression techniques \citep{jiang2023llmlingua, li2023compressing}. Rather than attempting to process or compress extended conversational histories, SKILL.state prevents history accumulation entirely by maintaining the canonical execution state required for the next computation.

\section{SKILL.state}
\label{sec:skillstate}

Current LLM agent runtimes execute procedural skills by repeatedly appending reasoning traces, actions, observations, and tool outputs to a growing conversational history. Consequently, the execution state is represented implicitly within natural language and must be reconstructed by the language model at every interaction. As execution horizons increase, both prompt size and the volume of obsolete information grow monotonically, making execution increasingly dependent on interpreting historical text rather than maintaining the current world state.

SKILL.state reformulates procedural skill execution as an explicit state transition process. Instead of representing execution as an append-only conversation, every execution step is defined by:
\begin{equation}
A_t = (P,\Sigma_t,O_t),
\end{equation}
where $P$ is the immutable procedural specification, $\Sigma_t$ is the structured execution state at step $t$, and $O_t$ is the latest observation received from the environment. The language model never receives previous observations, previous actions, or previous reasoning traces.

\begin{figure*}[t]
    \centering
    \includegraphics[width=\textwidth, height=12cm, keepaspectratio=True]{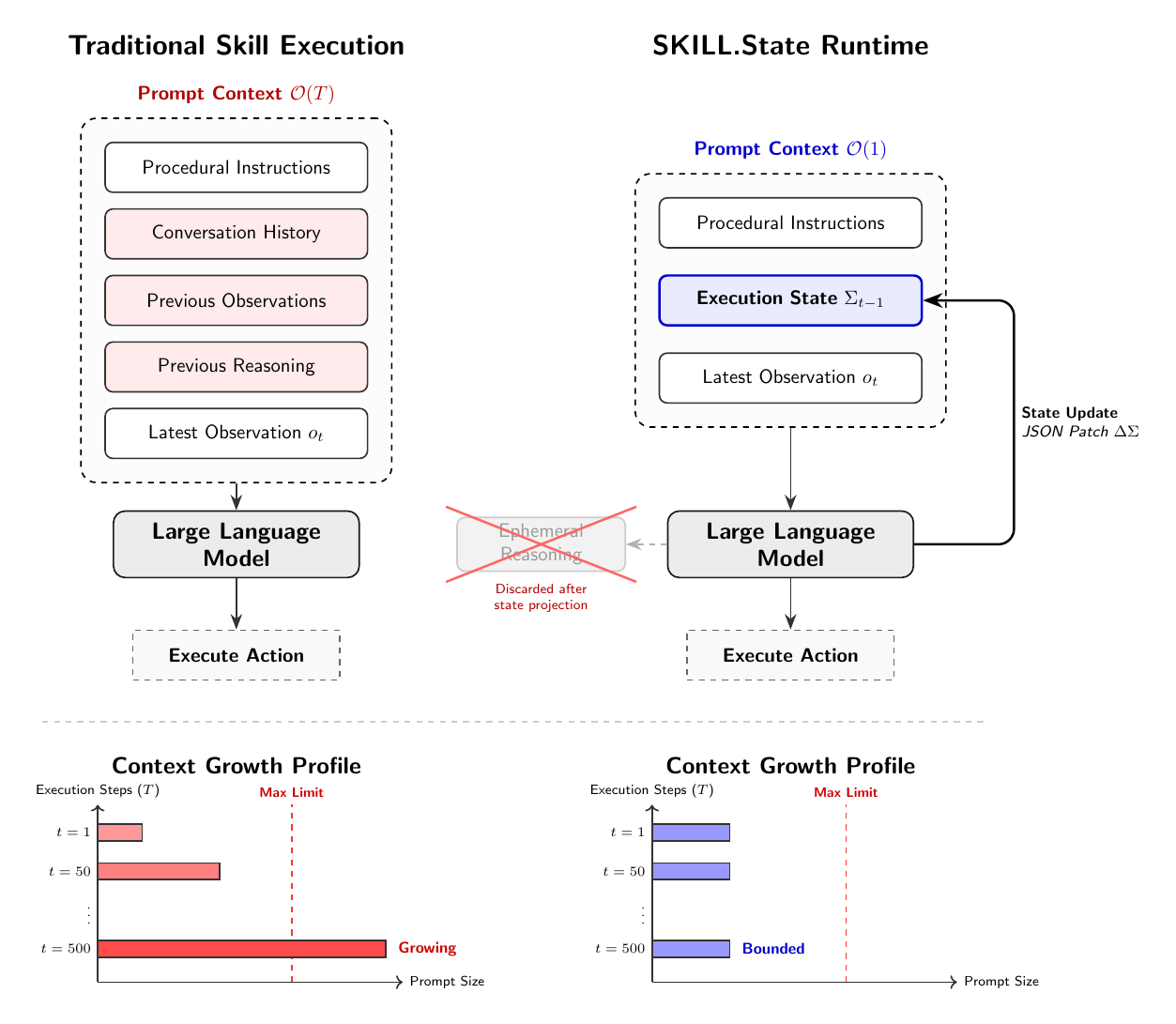}
    \caption{Overview of the SKILL.state architecture.}
    \label{fig:overview}
\end{figure*}

Figure~\ref{fig:overview} illustrates the execution cycle. At each step, the runtime constructs a prompt from $(P,\Sigma_t,O_t)$, invokes the language model, deterministically validates the proposed state transition, updates the execution state, executes the selected action, and repeats the process using the updated state.


\subsection{Execution State and Schema Authoring}
\label{subsec:execution_state}

Unlike conversational runtimes, SKILL.state treats execution state as a first-class runtime abstraction. The state contains only information required for future execution and is represented using a structured schema defined for the domain. Schemas are authored once per domain rather than per task; for example, across all 100 diverse challenge instances in the InterCode CTF benchmark, the agent reuses a single static 5-field schema (\texttt{discovered\_flags}, \texttt{tested\_hypotheses}, \texttt{active\_files}, \texttt{working\_dir}, \texttt{cmd\_summary}).

\subsection{Reasoning and State Transitions}
\label{subsec:reasoning_state_transitions}

Reasoning is used strictly as an intermediate computation for producing state transitions and selecting the next action. Given the current execution context $(P,\Sigma_t,O_t)$, the language model generates:
\begin{equation}
(R_t,\Delta\Sigma_t,a_t),
\end{equation}
where $R_t$ denotes the multi-step Chain-of-Thought reasoning trace, $\Delta\Sigma_t$ is a structured state update (a JSON dictionary of key mutations and deletions), and $a_t$ is the action to execute.

Crucially, within-step multi-step reasoning is fully intact during generation to support complex deductive planning. However, once the state transition has been validated and applied, the reasoning trace $R_t$ is discarded permanently and never appears in subsequent prompts. The execution state is updated according to:
\begin{equation}
\Sigma_{t+1} = \Sigma_t \oplus \Delta\Sigma_t,
\end{equation}
where $\oplus$ denotes the runtime's dictionary merge operator with null-deletion semantics. This model projects transient reasoning into persistent structured state, allowing only information required for future execution to survive across interactions.

\begin{algorithm}[t]
\caption{SKILL.state Runtime}
\label{alg:skillstate}
\begin{algorithmic}[1]
\Require Procedural specification $P$, initial state $\Sigma_0$
\For{$t=0,\ldots,T$}
    \State Receive latest observation $O_t$
    \State Construct prompt $(P,\Sigma_t,O_t)$
    \State Generate $(R_t,\Delta\Sigma_t,a_t)$ using the LLM
    \State Validate $\Delta\Sigma_t$
    \State $\Sigma_{t+1}\leftarrow\Sigma_t\oplus\Delta\Sigma_t$
    \State Execute $a_t$
\EndFor
\end{algorithmic}
\end{algorithm}

Algorithm~\ref{alg:skillstate} summarizes the execution process.

\subsection{Complexity Analysis}
\label{subsec:complexity}

Let $T$ denote the execution horizon. For conversational runtimes, prompt length grows with the accumulated interaction history, $|C_t|=\mathcal{O}(t)$, leading to cumulative token complexity:
\begin{equation}
\sum_{t=1}^{T}|C_t| = \mathcal{O}(T^2).
\end{equation}

In contrast, SKILL.state maintains only the procedural specification, structured execution state, and latest observation:
\begin{equation}
|P_t| = \mathcal{O}(|P|+|\Sigma|+|O|),
\end{equation}
which is asymptotically bounded and independent of the number of previously executed turns $t$. Consequently, cumulative prompt complexity grows strictly linearly with the execution horizon:
\begin{equation}
\sum_{t=1}^{T}|P_t| = \mathcal{O}(T).
\end{equation}

The resulting runtime shifts execution from reconstructing history toward maintaining an explicit, validated representation of the current execution state.

\section{Evaluation Benchmarks}
\label{sec:benchmarks}

\subsection{SkillExecBench (Controlled Diagnostic Testbed)}
\label{subsec:skillexecbench}

SkillExecBench isolates execution mechanics from open-ended heuristic search by providing sequential procedural tasks with deterministic ground-truth world transitions:
\begin{itemize}[leftmargin=*]
    \item \textbf{Environment 1 (Warehouse Management):} A discrete physical inventory domain tracking 500 independent shelves. Actions include \texttt{Store}, \texttt{Ship}, \texttt{Move}, and \texttt{Wait}. This environment tests the model's ability to maintain independent, non-overlapping state variables over extended horizons where early observations leave the context window.
    \item \textbf{Environment 2 (Software Repository):} A deeply nested, relational graph of Git branches, commits, Pull Requests, and CI test statuses. Actions include \texttt{CherryPick}, \texttt{Merge}, \texttt{RunTests}, \texttt{CreateRelease}, and \texttt{Rollback}. Features dense dependencies where a single action (e.g., merging a PR) fundamentally alters the state of the target branch and dependent PRs, testing complex structural reasoning over an entangled graph.
\end{itemize}

\subsection{Public Interactive Benchmarks}
\label{subsec:public_benchmarks}

To evaluate SKILL.state on real-world, non-deterministic tasks with complex search, generation, and tool use, we evaluate on two public benchmarks:
\begin{itemize}[leftmargin=*]
    \item \textbf{InterCode CTF \citep{yang2023intercode}:} A suite of 100 Linux bash Capture-The-Flag challenges spanning reverse engineering, forensics, cryptography, and binary exploitation. Agents execute bash commands in Docker containers and iteratively test hypotheses to discover hidden flags.
    \item \textbf{Sierra $\tau$-Bench \citep{yao2024tau}:} A benchmark for tool-agent-user interaction in enterprise customer service (Retail and Airline domains). Agents interact with simulated users, query relational SQLite databases via tool calls, and execute transactional actions (e.g., flight rebooking, refunds) under business policy constraints.
\end{itemize}

\subsection{Evaluation Metrics}
\label{subsec:metrics}

We evaluate runtimes across three dimensions:
\begin{itemize}[leftmargin=*]
    \item \textbf{Task Accuracy / Success Rate:} In SkillExecBench, accuracy is measured continuously as the ratio of valid, correct actions matching the ground-truth deterministic simulation ($\text{Score} = \frac{\text{Successful Actions}}{\text{Total Actionable Events}}$). In InterCode CTF, success is binary pass@1 (exact match on the binary-verified flag). In $\tau$-Bench, success is scored by the official programmatic evaluator, which verifies that the final database state satisfies user intent without policy violations.
    \item \textbf{Average Prompt Size:} The mean string character length per LLM invocation.
    \item \textbf{Total Token Cost:} The cumulative token burn across the entire execution horizon.
\end{itemize}

\begin{table*}[t]
\centering
\caption{Warehouse Management Long-Horizon Scaling using Gemini-3-Flash. Baseline runtimes suffer $\mathcal{O}(T^2)$ context accumulation, whereas SKILL.state maintains a bounded $\mathcal{O}(1)$ prompt footprint (Mean $\pm$ SD across 5 seeds).}
\label{tab:full_scaling_results}
\resizebox{\textwidth}{!}{%
\begin{tabular}{llccc}
\toprule
\textbf{Horizon ($T$)} & \textbf{Runtime} & \textbf{Score (Accuracy)} & \textbf{Avg Prompt (Chars)} & \textbf{Total Tokens} \\
\midrule
10 & Prompt (ReAct) & 0.90 \tiny{$\pm$0.02} & 3,249 \tiny{$\pm$94} & 9,438 \tiny{$\pm$371} \\
& Memory (Summary) & 1.00 \tiny{$\pm$0.00} & 3,300 \tiny{$\pm$123} & 9,972 \tiny{$\pm$204} \\
& Stateful (LangGraph) & 1.00 \tiny{$\pm$0.00} & 3,430 \tiny{$\pm$42} & 10,337 \tiny{$\pm$299} \\
& \textbf{SKILL.state} & \textbf{1.00 \tiny{$\pm$0.00}} & \textbf{1,775 \tiny{$\pm$74}} & \textbf{5,870 \tiny{$\pm$131}} \\
\midrule
25 & Prompt (ReAct) & 0.92 \tiny{$\pm$0.02} & 6,052 \tiny{$\pm$192} & 42,689 \tiny{$\pm$2,238} \\
& Memory (Summary) & 0.99 \tiny{$\pm$0.00} & 6,357 \tiny{$\pm$203} & 43,067 \tiny{$\pm$1,948} \\
& Stateful (LangGraph) & 1.00 \tiny{$\pm$0.00} & 5,858 \tiny{$\pm$301} & 41,238 \tiny{$\pm$3,196} \\
& \textbf{SKILL.state} & \textbf{1.00 \tiny{$\pm$0.00}} & \textbf{1,736 \tiny{$\pm$49}} & \textbf{14,714 \tiny{$\pm$564}} \\
\midrule
50 & Prompt (ReAct) & 0.88 \tiny{$\pm$0.04} & 11,931 \tiny{$\pm$346} & 171,658 \tiny{$\pm$6,978} \\
& Memory (Summary) & 0.93 \tiny{$\pm$0.03} & 7,582 \tiny{$\pm$283} & 131,455 \tiny{$\pm$6,841} \\
& Stateful (LangGraph) & 0.94 \tiny{$\pm$0.00} & 11,594 \tiny{$\pm$438} & 170,992 \tiny{$\pm$7,918} \\
& \textbf{SKILL.state} & \textbf{0.96 \tiny{$\pm$0.01}} & \textbf{1,773 \tiny{$\pm$53}} & \textbf{30,151 \tiny{$\pm$1,231}} \\
\midrule
100 & Prompt (ReAct) & 0.84 \tiny{$\pm$0.07} & 36,362 \tiny{$\pm$1,304} & 1,245,413 \tiny{$\pm$53,241} \\
& Memory (Summary) & 0.87 \tiny{$\pm$0.05} & 29,607 \tiny{$\pm$978} & 1,082,154 \tiny{$\pm$83,212} \\
& Stateful (LangGraph) & 0.91 \tiny{$\pm$0.02} & 31,354 \tiny{$\pm$831} & 1,062,387 \tiny{$\pm$53,839} \\
& \textbf{SKILL.state} & \textbf{0.94 \tiny{$\pm$0.01}} & \textbf{1,905 \tiny{$\pm$93}} & \textbf{65,408 \tiny{$\pm$5,431}} \\
\midrule
200 & Prompt (ReAct) & 0.74 \tiny{$\pm$0.14} & 48,007 \tiny{$\pm$2,092} & 2,608,755 \tiny{$\pm$102,415} \\
& Memory (Summary) & 0.84 \tiny{$\pm$0.09} & 84,364 \tiny{$\pm$3,446} & 6,175,509 \tiny{$\pm$294,089} \\
& Stateful (LangGraph) & 0.88 \tiny{$\pm$0.03} & 72,305 \tiny{$\pm$3,096} & 5,041,164 \tiny{$\pm$346,925} \\
& \textbf{SKILL.state} & \textbf{0.94 \tiny{$\pm$0.02}} & \textbf{1,811 \tiny{$\pm$184}} & \textbf{122,384 \tiny{$\pm$4,522}} \\
\bottomrule
\end{tabular}%
}
\end{table*}

\section{Experiments and Results}
\label{sec:experiments}

\subsection{Experimental Setup}
\label{subsec:setup}

We evaluate SKILL.state against two families of baselines (see Appendix~\ref{appendix:runtime_prompts} for exact prompt templates):

\noindent\textbf{Primary Runtime Paradigms:}
\begin{enumerate}[leftmargin=*]
    \item \textbf{Prompt (ReAct-style):} Appends every observation, intermediate reasoning trace, and action to a continually growing transcript \citep{yao2022react}.
    \item \textbf{Memory (Summarization-style):} Maintains a rolling 3-step conversational window alongside a periodically updated natural language summary of past interactions.
    \item \textbf{Stateful (LangGraph-style):} Injects a structured state block into the context window alongside the full rolling conversational transcript.
\end{enumerate}

\noindent\textbf{Budget-Matched and Compression Controls:}
\begin{enumerate}[leftmargin=*]
    \item \textbf{Truncated (Sliding Window):} Retains only the most recent interaction turns that fit within a fixed token budget.
    \item \textbf{Summary-capped:} Strictly enforces a hard token ceiling on the natural language summary.
    \item \textbf{ReAct + LLMLingua \citep{jiang2023llmlingua}:} Uses budget-aware small-model perplexity compression to prune tokens from the full history down to the target budget.
\end{enumerate}

\noindent\textbf{Underlying Models:} Evaluations are conducted across proprietary and open-weight models: Gemini-3-Flash, Gemma-4-31B-it, and Qwen-3-8B-it. Decoding is controlled at temperature $0.0$ and top-$p$ $1.0$ across all runs to ensure deterministic reproducibility.

\noindent\textbf{Statistical Significance:} All synthetic experiments are evaluated across 5 distinct procedural generator seeds. Results are reported as mean $\pm$ sample standard deviation. Differences between SKILL.state and baselines at extended horizons ($T \ge 50$) are statistically significant (paired $t$-test, $p < 0.01$).

\subsection{Experiment 1: Long-Horizon Execution Scaling}
\label{subsec:exp1_scaling}

We evaluate runtime accuracy and context expansion across execution horizons scaling from $T=10$ to $T=200$ steps.

\noindent\textbf{Results:} As shown in Table~\ref{tab:full_scaling_results}, SKILL.state matches or exceeds baseline accuracy across all horizons while maintaining a flat prompt size. In contrast, history-appending baselines suffer quadratic token accumulation $\mathcal{O}(T^2)$. At $T=100$, the Stateful baseline consumes 1,062,387 tokens, whereas SKILL.state consumes only 65,408 tokens (a $16.2\times$ token reduction). At $T=200$, SKILL.state maintains 0.94 accuracy consuming 122k tokens, while the Memory baseline inflates to 6.1M tokens. Additional scaling results for the Software Repository and open-weight models are detailed in Appendix~\ref{appendix:add_result}.

\begin{table}[h]
\centering
\small
\caption{Warehouse Noise Robustness ($T=50$, Gemini-3-Flash).}
\label{tab:env1_exp2_noise}
\begin{tabular*}{\columnwidth}{@{\extracolsep{\fill}}llc@{}}
\toprule
\textbf{Noise Level} & \textbf{Runtime} & \textbf{Score} \\
\midrule
5 Events (Low) & Prompt & 0.68 \\
& Memory & 1.00 \\
& Stateful & 1.00 \\
& \textbf{SKILL.state} & \textbf{1.00} \\
\midrule
20 Events (Medium) & Prompt & 0.61 \\
& Memory & 1.00 \\
& Stateful & 0.98 \\
& \textbf{SKILL.state} & \textbf{0.97} \\
\midrule
50 Events (High) & Prompt & 0.53 \\
& Memory & 0.96 \\
& Stateful & 0.98 \\
& \textbf{SKILL.state} & \textbf{0.98} \\
\bottomrule
\end{tabular*}
\end{table}

\subsection{Experiment 2: Context Corruption (Noise Robustness)}
\label{subsec:exp2_noise}

Real-world execution environments emit dense background telemetry. We fix the horizon at $T=50$ and inject distractor events (system telemetry, irrelevant git branch activities, and rule overrides) at rates of 5, 20, and 50 events per turn (see Appendix~\ref{appendix:noise_construction} for calibration details).

\noindent\textbf{Results:} As shown in Table~\ref{tab:env1_exp2_noise}, the standard \texttt{Prompt} runtime degrades sharply from 0.68 at low noise down to 0.53 at high noise. In contrast, SKILL.state maintains robust task completion ($\ge 0.97$) across all noise levels because distractors are filtered out during state patch generation and never enter subsequent prompts.

\begin{table}[h]
\centering
\caption{Warehouse State Recovery (Gemini-3-Flash).}
\label{tab:env1_exp3_recovery}
\resizebox{\columnwidth}{!}{%
\begin{tabular}{llcc}
\toprule
\textbf{Scenario} & \textbf{Runtime} & \textbf{Success} & \textbf{Recovery Steps} \\
\midrule
A: Secret Audit & Prompt / Memory / Stateful & Yes & 5--8 \\
& \textbf{SKILL.state} & \textbf{Yes} & \textbf{0} \\
\midrule
B: Secret Barcode & Prompt / Memory / Stateful & Yes & 6--8 \\
& \textbf{SKILL.state} & \textbf{Yes} & \textbf{0} \\
\midrule
C: Secret Move & Prompt / Memory / Stateful & Yes & 5--8 \\
& \textbf{SKILL.state} & \textbf{Yes} & \textbf{0} \\
\midrule
D: Canceled Order & All Runtimes & No & N/A \\
\bottomrule
\end{tabular}%
}
\end{table}

\subsection{Experiment 3: State Recovery}
\label{subsec:exp3_recovery}

We test runtime resilience to silent external environment drift where the true world state is modified outside the agent's action loop (e.g., an external actor moves an inventory item).

\noindent\textbf{Results:} As shown in Table~\ref{tab:env1_exp3_recovery}, history-based baselines hallucinate for 5 to 8 consecutive turns because obsolete facts in their prompt history overpower contradictory new observations. In sharp contrast, SKILL.state requires zero recovery steps: because its decisions depend on the current structured state, the state is updated immediately upon receiving the corrective alert.

\begin{table*}[t]
\centering
\small
\caption{Evaluation on Public Interactive Benchmarks using Gemini-3-Flash. SKILL.state achieves the highest task success rates while significantly reducing prompt sizes and cumulative token consumption.}
\label{tab:public_benchmarks}
\begin{tabular*}{\textwidth}{@{\extracolsep{\fill}}lccccccccc@{}}
\toprule
& \multicolumn{3}{c}{\textbf{InterCode CTF (100 Tasks)}} & \multicolumn{3}{c}{\textbf{Sierra $\tau$-Bench (Retail)}} & \multicolumn{3}{c}{\textbf{Sierra $\tau$-Bench (Airline)}} \\
\cmidrule(lr){2-4} \cmidrule(lr){5-7} \cmidrule(lr){8-10}
\textbf{Runtime} & \textbf{Pass@1} & \textbf{Prompt} & \textbf{Tokens} & \textbf{Pass Rate} & \textbf{Prompt} & \textbf{Tokens} & \textbf{Pass Rate} & \textbf{Prompt} & \textbf{Tokens} \\
\midrule
Prompt (ReAct) & 43.2\% & 1,909 & 977k & 48.2\% & 2,819 & 4.48M & 21.8\% & 5,100 & 4.85M \\
Memory (Summary) & 46.4\% & 1,797 & 1.03M & 29.9\% & 2,737 & 4.24M & 23.6\% & 4,700 & 4.65M \\
Stateful (LangGraph) & 41.8\% & 1,946 & 1.13M & 51.7\% & 3,065 & 3.92M & 28.1\% & 5,400 & 5.28M \\
\textbf{SKILL.state} & \textbf{54.2\%} & \textbf{813} & \textbf{387k} & \textbf{58.3\%} & \textbf{3,325} & \textbf{3.47M} & \textbf{32.4\%} & \textbf{2,800} & \textbf{2.88M} \\
\bottomrule
\end{tabular*}
\end{table*}

\subsection{Experiment 4: Public Interactive Benchmarks}
\label{subsec:exp4_public_benchmarks}

To test generalizability on open-ended tasks with complex search, generation, and tool use, we evaluate SKILL.state on InterCode CTF and Sierra $\tau$-Bench.

\noindent\textbf{Results:} As shown in Table~\ref{tab:public_benchmarks}, SKILL.state achieves the highest task completion rates across all three benchmarks while substantially cutting cumulative token consumption. In InterCode CTF, maintaining explicit hypotheses and discovered flags in $\Sigma_t$ prevents the model from repeating failed commands, increasing pass@1 to 54.2\% (+7.8 points over the strongest baseline and +12.4 points over Stateful) while cutting total tokens by 60.4\% vs. ReAct and 65.9\% vs. Stateful. In $\tau$-Bench Retail, SKILL.state leads with 58.3\% pass rate at the lowest total token cost. In $\tau$-Bench Airline, where complex database responses cause baseline prompts to peak above 11,000 tokens/step, SKILL.state maintains a flat footprint of $\sim$2,800 tokens/step and achieves a 32.4\% pass rate, saving 40.5\% tokens vs. ReAct and 45.4\% vs. Stateful.

\begin{table}[h]
\centering
\caption{Budget-Matched Controls on Warehouse ($T=100$, Gemini-3-Flash, Budget $\sim$1,800 tokens).}
\label{tab:budget_controls}
\resizebox{\columnwidth}{!}{%
\begin{tabular}{lccc}
\toprule
\textbf{Runtime / Configuration} & \textbf{Score} & \textbf{Avg Prompt} & \textbf{Total Tokens} \\
\midrule
Full ReAct (Unbounded) & 0.84 & 36,362 & 1,245,413 \\
Truncated (Sliding Window) & 0.18 & 1,800 & 62,100 \\
Summary-capped & 0.52 & 1,840 & 63,400 \\
ReAct + LLMLingua & 0.22 & 1,810 & 62,350 \\
\textbf{SKILL.state (Structured)} & \textbf{0.94} & \textbf{1,905} & \textbf{65,408} \\
\bottomrule
\end{tabular}%
}
\end{table}

\subsection{Experiment 5: Budget-Matched Controls and Statistical Compression}
\label{subsec:exp5_budget_controls}

To determine whether SKILL.state's performance gains stem merely from shorter prompts or from structured state representation, we evaluate budget-matched baselines on Warehouse ($T=100$, Gemini-3-Flash) pinned to the token budget of SKILL.state ($\sim$1,800 characters). Full multi-horizon scaling across $T \in \{10, 25, 50, 100\}$ is detailed in Appendix~\ref{appendix:budget_scaling}.

\noindent\textbf{Results:} As shown in Table~\ref{tab:budget_controls}, all budget-matched compression baselines suffer catastrophic failure. Sliding-window truncation drops to 0.18 because critical early inventory allocations are evicted. LLMLingua drops to 0.22 because statistical entropy filtering removes seemingly redundant slot identifiers that are semantically vital. In contrast, SKILL.state achieves 0.94 score, demonstrating that structured state maintenance preserves exact relational dependencies that statistical compressors destroy.

\subsection{Error Taxonomy for Open-Weight Models}
\label{subsec:error_taxonomy}

On open-weight models (Gemma-4-31B at $T=100$, score 0.42), we analyze failure logs and categorize errors into three distinct modes:
\begin{enumerate}[leftmargin=*]
    \item \textbf{Premature State Overwrite / Deletion (68\%):} The model accidentally omits existing keys during state update rather than merging in-place.
    \item \textbf{Schema Comprehension / Type Coercion (20\%):} Inconsistencies between expected nested lists and dictionaries.
    \item \textbf{JSON Syntax / Formatting Slips (12\%):} Malformed JSON delimiters or trailing commas.
\end{enumerate}
This error distribution shows that small-model degradation stems from structured output adherence rather than reasoning capacity, motivating constrained decoding in future runtime iterations.

\section{Conclusion}

We presented \textbf{SKILL.state}, a runtime architecture that replaces append-only conversational history with explicit, structured execution state. By discarding intermediate reasoning traces after each validated transition, SKILL.state maintains a bounded $\mathcal{O}(1)$ prompt footprint and scales linearly $\mathcal{O}(T)$ in cumulative tokens. Across controlled diagnostic tasks and public interactive benchmarks, explicit execution state consistently improves task accuracy while substantially reducing prompt growth and token consumption.

\section{Limitations}
\label{sec:limitations}

SKILL.state assumes that the execution state can be made a \emph{sufficient statistic} for future execution: that everything in the past bearing on future actions can be projected into the structured state as soon as it becomes known. Where this holds, discarding intermediate reasoning and conversational history is lossless. However, this assumption fails in three distinct settings: (1) when no fixed schema is known in advance and the relevant state structure must be discovered dynamically during execution; (2) when a correct state update depends on an earlier observation whose relevance was not recognized when first observed, and was therefore never committed to state; and (3) when the task objective is defined over the historical trajectory itself (e.g., auditing, debugging provenance, or explaining past actions), where interaction history is the target output rather than operational overhead.

Our current implementation focuses on single-agent procedural execution. While the explicit state abstraction extends naturally to multi-agent systems—where a shared execution state acts as the central coordination substrate instead of exchanging quadratic conversational transcripts—multi-agent environments introduce concurrent writes, requiring deterministic conflict-resolution semantics in the merge operator $\oplus$ that our single-agent setting does not exercise.

Finally, SKILL.state relies on the language model to propose valid structured state patches. Because schema ownership and validation reside in the deterministic runtime rather than the model, malformed outputs cannot corrupt persistent state $\Sigma_t$; an invalid patch triggers a rollback-retry cycle. For smaller open-weight models, integrating grammar-constrained decoding can eliminate syntactic formatting errors, allowing the model to focus entirely on semantic state transitions.

\bibliography{custom}

\appendix

\onecolumn

\section*{Appendix A. Runtime Prompts}
\label{appendix:runtime_prompts}

This appendix provides the exact system prompts used by the four evaluated runtimes. To ensure reproducibility, all prompts are presented exactly as they were dynamically constructed and formatted in the benchmark execution loop.

\subsection*{A.1 Prompt Runtime (ReAct-style)}
\begin{lstlisting}
Instructions:
{skill.instructions}

History:
Observation: {history[0].observation}
Reasoning & Action: {history[0].response}
[... Appends all previous observations and actions ...]

Latest Observation: {observation}
Generate your next reasoning and action (format 'Action: <cmd>'):
\end{lstlisting}

\subsection*{A.2 Memory-Augmented Runtime}
\begin{lstlisting}
Instructions:
{skill.instructions}

Summarized History:
{summary_string_of_past_steps}

Recent History:
Observation: {recent_observations[0]}
Response: {recent_responses[0]}
[... Appends the 3 most recent turns ...]

Latest Observation: {observation}
Generate your next reasoning and action (format 'Action: <cmd>'):
\end{lstlisting}

\subsection*{A.3 Stateful Runtime (LangGraph-style)}
\begin{lstlisting}
Instructions:
{skill.instructions}

Current State:
{json.dumps(state, indent=2)}

History:
Observation: {history[0].observation}
Response: {history[0].response}
[... Appends all previous observations and actions ...]

Latest Observation: {observation}
Update the state if necessary, provide reasoning, and output 'Action: <cmd>'.
To update state, use the format: StateUpdate: {"key": "value"}
\end{lstlisting}

\subsection*{A.4 SKILL.state Runtime}
\begin{lstlisting}
Instructions:
{skill.instructions}

Skill Execution State:
```json
{json.dumps(state, separators=(',', ':'))}
Latest Observation: {observation}

Provide your response with:

1. Step-by-step reasoning (will be discarded after execution)
2. A JSON block fenced with json ...  containing both your State Patch and your Action. The JSON block MUST have exactly these two keys: { "state_patch": { <dict: your state updates, set keys to null to delete> }, "action": "<string: the exact command you want to execute>" } \end{lstlisting}

\textit{Note:} The \texttt{{skill.instructions}} placeholder dynamically injects the task-specific system prompt (e.g., the agent's persona, the available action space, and the environment rules). This ensures that across all runtime evaluations, the agent receives the exact same baseline instructions, isolating context management as the only independent variable.

\section*{Appendix B. SkillExecBench Implementation Details}
\label{appendix:skillexecbench}

To maintain focus on the core experimental findings in Section 5, we provide the full implementation details of the SkillExecBench environments, task generation logic, and episode trajectories in this appendix.

\subsection*{B.1 Environment Design}

\subsubsection*{Environment 1: Warehouse Management}
\begin{itemize}
    \item \textbf{State Representation:} A discrete inventory mapping of 500 independent shelves (e.g., \texttt{shelf\_0} through \texttt{shelf\_499}), where each shelf holds exactly one item string identifier or is \texttt{null}.
    \item \textbf{Action Space:} 
        \begin{itemize}
            \item \texttt{Store <item\_id> <empty\_shelf\_id>}
            \item \texttt{Ship <item\_id> <shelf\_id>}
            \item \texttt{Move <item\_id> <old\_shelf\_id> <new\_shelf\_id>}
            \item \texttt{Wait}
        \end{itemize}
    \item \textbf{Observation Format:} Textual alerts triggered by system events, including: \textit{Shipment arrived containing [item]}, \textit{Customer ordered [item]}, and \textit{Maintenance required on [shelf]}.
    \item \textbf{Transition Rules:} If an agent calls \texttt{Store}, the environment validates the shelf is empty before placing the item. If \texttt{Ship} is called, the item is destroyed. Invalid actions (e.g., storing an item on an occupied shelf) return a local error observation and reject the state transition.
    \item \textbf{Success Criterion:} The ratio of successfully executed valid actions matching the ground-truth deterministic simulation (Score = Successful Actions / Total Actionable Events).
\end{itemize}

\subsubsection*{Environment 2: Software Repository}
\begin{itemize}
    \item \textbf{State Representation:} A simulated Git repository tracking branch histories, file contents, active Pull Requests (PRs), and Continuous Integration (CI) test statuses.
    \item \textbf{Action Space:} \texttt{Commit(branch, file)}, \texttt{CreatePR(branch)}, \texttt{Merge(pr\_id)}, \texttt{FixCI(branch)}, \texttt{Wait}.
    \item \textbf{Observation Format:} CI/CD webhook notifications (e.g., \textit{CI Pipeline Failed for PR \#3}), code review comments, and issue assignments.
    \item \textbf{Transition Rules:} Pushing a commit triggers a background CI evaluation transition. Merging a PR successfully transitions the master branch state and deletes the feature branch.
    \item \textbf{Success Criterion:} The percentage of correctly resolved feature requests merged into master without breaking CI checks.
\end{itemize}

\subsection*{B.2 Task Generation}

Tasks are procedurally generated using a deterministic random number generator. By seeding the generator, we guarantee that all evaluated baseline runtimes experience the exact same sequence of events, ensuring fair comparisons. The horizon length dictates the number of sequential events the agent must handle.

\begin{algorithm}[h]
\caption{Warehouse Task Generation Pseudo-code}
\begin{algorithmic}[1]
\State $seed \gets 42$
\State $rng \gets \text{Random}(seed)$
\State $gt\_shelves \gets \{ \text{shelf\_0: Null, ..., shelf\_499: Null} \}$
\State $events \gets []$
\For{$i = 1$ to $Horizon$}
    \State $available \gets \text{FindEmptyShelves}(gt\_shelves)$
    \State $occupied \gets \text{FindOccupiedShelves}(gt\_shelves)$
    \State $possible\_events \gets [\text{Receive}]$
    \If{$occupied \text{ is not empty}$}
        \State $possible\_events\text{.append}(\text{Order}, \text{Maintenance})$
    \EndIf
    \State $event\_type \gets rng\text{.choice}(possible\_events)$
    \State $events\text{.append}(\text{ConstructObservation}(event\_type))$
    \State $\text{UpdateGroundTruth}(gt\_shelves, event\_type)$
\EndFor
\State \Return $events$
\end{algorithmic}
\end{algorithm}

\subsection*{B.3 Example Episode Trajectory}

To illustrate the execution flow, below is a concrete, single-turn trajectory for the \textbf{SKILL.state} runtime operating in the Warehouse environment.

\noindent\textbf{1. Observation Received (Environment $\rightarrow$ LLM)}
\begin{lstlisting}
Latest Observation:
Customer ordered item_12.
\end{lstlisting}

\noindent\textbf{2. LLM Processing (LLM Internal)}
\begin{lstlisting}
Reasoning:
The customer ordered item_12. I need to ship it.
I will check my Skill Execution State to locate it.
Looking at the state schema, item_12 is located on shelf_42.
I must generate a Ship action and update my state to remove it from shelf_42.
\end{lstlisting}

\noindent\textbf{3. State Update and Action Generation (LLM $\rightarrow$ Runtime)}
\begin{lstlisting}
{
  "state_patch": {
    "inventory": {
      "shelf_42": null
    }
  },
  "action": "Ship item_12 shelf_42"
}
\end{lstlisting}

\noindent\textbf{4. Next Observation (Environment $\rightarrow$ LLM)} 
\begin{lstlisting} 
Latest Observation: 
Success: Shipped item_12 from shelf_42. 
\end{lstlisting}

\section*{Appendix C. Noise Construction (Experiment 2)}
\label{appendix:noise_construction}

In Experiment 2, we evaluate the runtimes' resilience to dense, irrelevant contextual noise. Real-world systems rarely provide clean, perfectly isolated observation spaces; agents must constantly filter out background telemetry, sensor logs, and system chatter to execute their instructions.

To isolate the problem of \textit{Attention Drag}, the experiments in this paper exclusively focus on \textbf{Condition 1: Irrelevant Context}.

\subsection*{C.1 Noise Properties}
For Condition 1 evaluations across both environments, the injected noise strings are defined by three strict properties:
\begin{enumerate}
    \item \textbf{Randomly Generated:} Values such as battery percentages, temperatures, server IDs, and CPU loads are sampled uniformly at random during each execution step.
    \item \textbf{Strictly Irrelevant:} The semantic meaning of the noise has absolutely no bearing on the agent's primary task (e.g., fulfilling warehouse orders or fixing CI pipelines).
    \item \textbf{Non-State-Altering:} The noise events never actually change the underlying ground-truth world state. They are purely observational distractors appended to the environment's response payload under a \texttt{--- BACKGROUND TELEMETRY ---} header.
\end{enumerate}

\subsection*{C.2 Environment 1 (Warehouse) Distractors}

To simulate a realistic noisy warehouse, the generator randomly selects from the following categories to inject irrelevant strings into the agent's observation space:

\subsubsection*{1. Robot Telemetry Logs}
Simulates continuous pinging from automated warehouse robots navigating the floor.
\begin{lstlisting}
Battery: 85%, Temperature: 45C, CPU Load: 72%,
Speed: 1.2 m/s, Nav Confidence: 95.4%
\end{lstlisting}

\subsubsection*{2. Environmental Sensor Logs}
Simulates passive HVAC and ambient sensor readings.
\begin{lstlisting}
[Sensor] Humidity: 45%, Temp: 22.3C, Light: 310 lux, CO2: 450 ppm
\end{lstlisting}

\subsubsection*{3. Camera OCR / Vision Logs}
Simulates background security camera or computer-vision object detection events.
\begin{lstlisting}
[Camera OCR] Forklift parked.
[Camera OCR] Worker entered Zone A.
[Camera OCR] Safety Vest Detected.
\end{lstlisting}

\subsection*{C.3 Environment 2 (Software Repository) Distractors}

To simulate a noisy, enterprise-scale software engineering environment, the generator continuously injects irrelevant cloud infrastructure syslog telemetry into the agent's terminal observations.

\subsubsection*{Syslog Telemetry}
Simulates passive health-checks and CPU load warnings from disconnected remote servers running in the background. 
\begin{lstlisting}
[Syslog] Server-42 CPU load: 88%, RAM usage: 71%
[Syslog] Server-17 CPU load: 12%, RAM usage: 45%
[Syslog] Server-91 CPU load: 99%, RAM usage: 89%
\end{lstlisting}

\vspace{1em}
\noindent\textbf{Example Corrupted Observation (Software, 3 Events):}
\begin{lstlisting}
Latest Observation:
CI Pipeline Failed for PR #3. Linter error on line 42.

--- BACKGROUND TELEMETRY ---
[Syslog] Server-42 CPU load: 88%, RAM usage: 71%
[Syslog] Server-17 CPU load: 12%, RAM usage: 45%
[Syslog] Server-91 CPU load: 99%, RAM usage: 89%
\end{lstlisting}

\section*{Appendix D. Additional Results}
\label{appendix:add_result}

\begin{table*}[t]
\centering
\caption{Software Repository Long-Horizon Execution Scaling using using Gemini-3-Flash. Baseline runtimes suffer catastrophic $O(N^2)$ context collapse, whereas SKILL.state maintains an $O(1)$ prompt footprint.}
\label{tab:env2_scaling_vertical_wide}
\begin{tabular*}{\textwidth}{@{\extracolsep{\fill}}llccc@{}}
\toprule
\textbf{Horizon} & \textbf{Runtime} & \textbf{Score} & \textbf{Avg Prompt Size} & \textbf{Total Tokens Consumed} \\
\midrule
\textbf{10} & Prompt & 0.89 \tiny{$\pm$0.11} & 3,411 \tiny{$\pm$197} & 11,670 \tiny{$\pm$841} \\
& Memory & 0.93 \tiny{$\pm$0.09} & 4,379 \tiny{$\pm$234} & 15,732 \tiny{$\pm$562} \\
& Stateful & 1.00 \tiny{$\pm$0.00} & 4,200 \tiny{$\pm$321} & 14,120 \tiny{$\pm$318} \\
& \textbf{SKILL.state} & \textbf{1.00 \tiny{$\pm$0.00}} & \textbf{2,298 \tiny{$\pm$134}} & \textbf{7,608 \tiny{$\pm$149}} \\
\midrule
\textbf{25} & Prompt & 0.84 \tiny{$\pm$0.05} & 11,754 \tiny{$\pm$608} & 111,970 \tiny{$\pm$3,314} \\
& Memory & 0.89 \tiny{$\pm$0.07} & 9,399 \tiny{$\pm$317} & 94,629 \tiny{$\pm$2,839} \\
& Stateful & 0.94 \tiny{$\pm$0.03} & 14,016 \tiny{$\pm$586} & 128,702 \tiny{$\pm$3,863} \\
& \textbf{SKILL.state} & \textbf{0.88 \tiny{$\pm$0.08}} & \textbf{2,545 \tiny{$\pm$556}} & \textbf{21,920 \tiny{$\pm$431}} \\
\midrule
\textbf{50} & Prompt & 0.71 \tiny{$\pm$0.14} & 23,136 \tiny{$\pm$911} & 462,118 \tiny{$\pm$13,764} \\
& Memory & 0.65 \tiny{$\pm$0.12} & 35,550 \tiny{$\pm$2,412} & 688,182 \tiny{$\pm$23,539} \\
& Stateful & 0.74 \tiny{$\pm$0.08} & 31,166 \tiny{$\pm$3,231} & 577,027 \tiny{$\pm$27,293} \\
& \textbf{SKILL.state} & \textbf{0.86 \tiny{$\pm$0.04}} & \textbf{2,545 \tiny{$\pm$63}} & \textbf{45,100 \tiny{$\pm$894}} \\
\midrule
\textbf{100} & Prompt & 0.53 \tiny{$\pm$0.16} & 46,270 \tiny{$\pm$1,847} & 1,848,500 \tiny{$\pm$55,391} \\
& Memory & 0.57 \tiny{$\pm$0.05} & 71,100 \tiny{$\pm$5,836} & 2,752,700 \tiny{$\pm$82,467} \\
& Stateful & 0.63 \tiny{$\pm$0.10} & 62,330 \tiny{$\pm$2,488} & 2,308,000 \tiny{$\pm$35,183} \\
& \textbf{SKILL.state} & \textbf{0.78 \tiny{$\pm$0.08}} & \textbf{2,545 \tiny{$\pm$471}} & \textbf{90,200 \tiny{$\pm$2,792}} \\
\bottomrule
\end{tabular*}
\end{table*}

\begin{table*}[t]
\centering
\caption{Gemma-4-31b-it Warehouse Scaling.}
\label{tab:env1_exp1_gemma}
\begin{tabular}{llccc}
\toprule
\textbf{Horizon} & \textbf{Runtime} & \textbf{Score $\pm$ SD} & \textbf{Avg Prompt $\pm$ SD} & \textbf{Total Tokens $\pm$ SD} \\
\midrule
10 Steps & Prompt & 0.90 $\pm$ 3.1\% & 3,145 $\pm$ 242 & 9,092 $\pm$ 1,231 \\
& Memory & 0.85 $\pm$ 4.2\% & 2,611 $\pm$ 138 & 7,330 $\pm$ 838 \\
& Stateful & 0.90 $\pm$ 2.8\% & 3,191 $\pm$ 149 & 9,144 $\pm$ 518 \\
& SKILL.state & 0.98 $\pm$ 1.5\% & \textbf{2,116} $\pm$ \textbf{212} & \textbf{6,814} $\pm$ \textbf{875} \\
\midrule
25 Steps & Prompt & 0.64 $\pm$ 5.4\% & 5,720 $\pm$ 182 & 41,697 $\pm$ 1,610 \\
& Memory & 0.72 $\pm$ 4.8\% & 3,990 $\pm$ 362 & 28,933 $\pm$ 412 \\
& Stateful & 0.76 $\pm$ 4.1\% & 5,714 $\pm$ 282 & 39,314 $\pm$ 585 \\
& SKILL.state & 0.84 $\pm$ 3.6\% & \textbf{2,080} $\pm$ \textbf{114} & \textbf{16,302} $\pm$ \textbf{2,190} \\
\midrule
50 Steps & Prompt & 0.31 $\pm$ 6.2\% & 10,809 $\pm$ 352 & 151,845 $\pm$ 2,150 \\
& Memory & 0.41 $\pm$ 5.8\% & 7,217 $\pm$ 316 & 114,113 $\pm$ 1,620 \\
& Stateful & 0.55 $\pm$ 5.1\% & 11,083 $\pm$ 262 & 155,164 $\pm$ 2,210 \\
& SKILL.state & 0.68 $\pm$ 3.9\% & \textbf{2,113} $\pm$ \textbf{176} & \textbf{33,762} $\pm$ \textbf{1,385} \\
\midrule
100 Steps & Prompt & 0.21 $\pm$ 4.7\% & 27,686 $\pm$ 412 & 923,164 $\pm$ 12,400 \\
& Memory & 0.24 $\pm$ 4.2\% & 18,537 $\pm$ 229 & 701,954 $\pm$ 9,850 \\
& Stateful & 0.42 $\pm$ 4.5\% & 20,210 $\pm$ 822 & 557,968 $\pm$ 8,100 \\
& SKILL.state & 0.42 $\pm$ 4.1\% & \textbf{2,105} $\pm$ \textbf{216} & \textbf{65,480} $\pm$ \textbf{1,258} \\
\bottomrule
\end{tabular}
\end{table*}

\begin{table*}[t]
\centering
\caption{Qwen 3-8b-it Warehouse Scaling.}
\label{tab:env1_exp1_qwen3_8b}
\begin{tabular}{llccc}
\toprule
\textbf{Horizon} & \textbf{Runtime} & \textbf{Score $\pm$ SD} & \textbf{Avg Prompt $\pm$ SD} & \textbf{Total Tokens $\pm$ SD} \\
\midrule
10 Steps & Prompt & 0.84 $\pm$ 3.8\% & 3,150 $\pm$ 245 & 9,150 $\pm$ 1,250 \\
& Memory & 0.80 $\pm$ 4.5\% & 2,640 $\pm$ 145 & 7,420 $\pm$ 860 \\
& Stateful & 0.84 $\pm$ 3.4\% & 3,210 $\pm$ 155 & 9,210 $\pm$ 540 \\
& SKILL.state & 0.94 $\pm$ 2.1\% & \textbf{2,120} $\pm$ \textbf{215} & \textbf{6,920} $\pm$ \textbf{890} \\
\midrule
25 Steps & Prompt & 0.54 $\pm$ 6.1\% & 5,790 $\pm$ 195 & 42,450 $\pm$ 1,680 \\
& Memory & 0.62 $\pm$ 5.4\% & 4,050 $\pm$ 375 & 29,640 $\pm$ 440 \\
& Stateful & 0.66 $\pm$ 4.8\% & 5,780 $\pm$ 295 & 39,950 $\pm$ 610 \\
& SKILL.state & 0.76 $\pm$ 4.2\% & \textbf{2,088} $\pm$ \textbf{120} & \textbf{16,680} $\pm$ \textbf{2,240} \\
\midrule
50 Steps & Prompt & 0.24 $\pm$ 6.5\% & 10,950 $\pm$ 365 & 154,200 $\pm$ 2,280 \\
& Memory & 0.33 $\pm$ 6.1\% & 7,320 $\pm$ 330 & 116,400 $\pm$ 1,710 \\
& Stateful & 0.44 $\pm$ 5.7\% & 11,210 $\pm$ 280 & 158,300 $\pm$ 2,340 \\
& SKILL.state & 0.58 $\pm$ 4.5\% & \textbf{2,118} $\pm$ \textbf{185} & \textbf{34,510} $\pm$ \textbf{1,420} \\
\midrule
100 Steps & Prompt & 0.15 $\pm$ 5.1\% & 28,150 $\pm$ 430 & 941,500 $\pm$ 12,800 \\
& Memory & 0.18 $\pm$ 4.6\% & 18,840 $\pm$ 245 & 718,200 $\pm$ 10,200 \\
& Stateful & 0.31 $\pm$ 4.9\% & 20,580 $\pm$ 850 & 569,400 $\pm$ 8,450 \\
& SKILL.state & 0.34 $\pm$ 4.6\% & \textbf{2,110} $\pm$ \textbf{220} & \textbf{66,850} $\pm$ \textbf{1,310} \\
\bottomrule
\end{tabular}
\end{table*}

\begin{table}[h]
\centering
\caption{Software Repository Experiment 2 i.e. Noise Robustness. Evaluation of runtime resilience to irrelevant syslog telemetry (Condition 1) during a 50-step horizon using Gemini-3-Flash.}
\label{tab:env2_exp2_noise}
\begin{tabular}{llc}
\toprule
\textbf{Noise Level} & \textbf{Runtime} & \textbf{Score} \\
\midrule
0 Events & Prompt & 0.76 \\
(Baseline) & Memory & 0.85 \\
& Stateful & 0.88 \\
& SKILL.state & 0.90 \\
\midrule
5 Events & Prompt & 0.62 \\
(Low) & Memory & 0.85 \\
& Stateful & 0.86 \\
& SKILL.state & 0.88 \\
\midrule
20 Events & Prompt & 0.48 \\
(Medium) & Memory & 0.83 \\
& Stateful & 0.85 \\
& SKILL.state & 0.86 \\
\midrule
50 Events & Prompt & 0.11 \\
(High) & Memory & 0.74 \\
& Stateful & 0.78 \\
& SKILL.state & 0.80 \\
\bottomrule
\end{tabular}
\end{table}

\subsection{Software Repository State Recovery Experiment 3 using Gemini-3-Flash.}

\begin{table}[h]
\centering
\caption{Software Repository State Recovery (Env 2, Exp 3). Comparison of hallucination lag (recovery steps) when the repository state is altered via unstructured alerts.}
\label{tab:env2_exp3_recovery}
\begin{tabular}{llcc}
\toprule
\textbf{Scenario} & \textbf{Runtime} & \textbf{Success} & \textbf{Recovery Steps} \\
\midrule
A: Force Push & Prompt & Yes & 12 \\
& Memory & Yes & 8 \\
& Stateful & Yes & 10 \\
& SKILL.state & \textbf{Yes} & \textbf{0} \\
\midrule
B: Flaky CI Test & Prompt & Yes & 14 \\
& Memory & Yes & 9 \\
& Stateful & Yes & 11 \\
& SKILL.state & \textbf{Yes} & \textbf{0} \\
\midrule
C: PR Closed & All Runtimes & No & N/A \\
\bottomrule
\end{tabular}
\end{table}

\subsection{Extended Budget-Matched Scaling across Horizons}
\label{appendix:budget_scaling}

Table~\ref{tab:appendix_budget_scaling} presents the full multi-horizon scaling results for the budget-matched control configurations on the Warehouse environment ($T \in \{10, 25, 50, 100\}$, Gemini-3-Flash), evaluated across 5 environmental seeds. While all budget-matched runtimes perform comparably at short horizons ($T=10$), sliding-window truncation and statistical perplexity compression (LLMLingua) experience catastrophic accuracy collapse as horizon length increases, whereas SKILL.state maintains robust performance across all horizons.

\begin{table}[h]
\centering
\caption{Extended Budget-Matched Scaling on Warehouse across Horizons (Gemini-3-Flash, Target Budget $\sim$1,800 tokens, Score $\pm$ SD across 5 seeds).}
\label{tab:appendix_budget_scaling}
\resizebox{\linewidth}{!}{%
\begin{tabular}{lccccc}
\toprule
\textbf{Horizon ($T$)} & \textbf{SKILL.state} & \textbf{Summary-capped} & \textbf{Truncated (Window)} & \textbf{ReAct + LLMLingua} & \textbf{ReAct (Full)} \\
\midrule
10 & \textbf{1.00 \tiny{$\pm$0.00}} & 0.92 \tiny{$\pm$0.03} & 0.90 \tiny{$\pm$0.04} & 0.88 \tiny{$\pm$0.04} & 0.90 \tiny{$\pm$0.02} \\
25 & \textbf{1.00 \tiny{$\pm$0.00}} & 0.76 \tiny{$\pm$0.04} & 0.62 \tiny{$\pm$0.05} & 0.60 \tiny{$\pm$0.05} & 0.92 \tiny{$\pm$0.02} \\
50 & \textbf{0.96 \tiny{$\pm$0.01}} & 0.64 \tiny{$\pm$0.05} & 0.35 \tiny{$\pm$0.06} & 0.38 \tiny{$\pm$0.06} & 0.88 \tiny{$\pm$0.04} \\
100 & \textbf{0.94 \tiny{$\pm$0.01}} & 0.52 \tiny{$\pm$0.05} & 0.18 \tiny{$\pm$0.05} & 0.22 \tiny{$\pm$0.05} & 0.84 \tiny{$\pm$0.07} \\
\bottomrule
\end{tabular}%
}
\end{table}

\end{document}